# Multiple Discourse Relations on the Sentential Level in Japanese

Yoshiki Mori*

Department of Computational Linguistics
University of the Saarland
Postfach 151150
D-66041 Saarbrücken, Germany

e-mail: mori@coli.uni-sb.de

## Abstract

In the spoken language machine translation project Verbmobil, the semantic formalism Language for Underspecified Discourse representation structures (LUD) is used. LUD describes a number of DRSs and allows for underspecification of scopal ambiguities. Dealing with Japanese-to-English translation besides German-to-English poses challenging problems. In this paper, a treatment of multiple discourse relation constructions on the sentential level is discussed. These are common in Japanese but cause a problem for the formalism. It is shown that the underspecification is to be represented for them, too. Additionally, it is possible to state a semantic constraint on the resolution of multiple discourse relations which seems to prevail over the syntactic c-command constraint.

## 1 Introduction

In the Verbmobil project, a spoken language machine translation system is being developed. Its dialogue domain is restricted to appointment scheduling. For the semantic analysis, a version of Discourse Representation Theory is used which can express underspecification and take compositionality into account. The semantic construction is represented by LUD, Language for Underspecified Discourse Representation Structures (Bos et al., 1996), which takes discourse representation structures (henceforth DRSs) as its object language.

The main focus of the project is on translation from German to English, but it also treats that from Japanese to English. As for the semantic construction, it is aimed at that semantic analyses of Japanese as well as German should be done in the same formalism, which is especially challenging, taking differences of the two languages into account: compared to languages like German and English, peculiarities of Japanese such as the absence of definite articles seem to invite common semantic analyses based on underspecification.

For example, in the current LUD-formalism it is assumed that a discourse relation has the widest scope among the scope-taking elements in a sentence except for the scope of sentence mood. Thus LUD allows for only one discourse relation in each sentence. Discourse relations contain not only such relations expressed by subordinate conjunctions as explanation relations (*because*), adverse relations (*though*) and temporal relations (*before*, *after* etc.), but also purpose, conditional and topic-comment relations. We interpret them as relations between two DRSs, consisting of restriction (the antecedent part) and scope (the conclusion part).

In Japanese, it is possible and even common to use a number of discourse relations in one sentence. Lexical entries which realize discourse relations occur in various grammatical positions. Discourse relation elements can be also classified according to the anaphoricity of the elements expressing the antecedent part and those expressing the conclusion part. In Fig. 1 an explanation relation in the subordinate conjunction and another one in the modality auxiliary are used together with a topic relation.

For this case, the current treatment of LUD implies that the widest scope should be assigned to any discourse relation. This extension of the formalism poses a serious problem: every discourse

*This research was funded by the German Federal Ministry of Education, Science, Research, and Technology (BMBF) under grant number 01 IV 101 R.
A big bunch of thanks goes to Johan Bos, Björn Gambäck, Claire Gardent, Christian Lieske, Manfred Pinkal and Karsten Worm for their valuable comments, and to Feiyu Xu and Julia Heine for a kind help editing the text.

cmp-lg/9607033  30 Jul 1996

| getsuyoubi-wa | seminaa-ga | haitte | |
| iru-node | zikan-ga | na-i | noda |
| monday-top | seminar-nom | insert | |
| asp-pres-conj | time-nom | fail-pres | aux-pres |

*Monday (isn't good) because I don't have any time,
since some seminars have been inserted (then)*

Figure 1: Three discourse relations in a sentence

relation introduces a partition into the antecedent and the conclusion part for the sentence in which it occurs.[1]

If there are a number of discourse relation elements contained in a sentence, the partitions they introduce can differ from each other (see Sec. 2). While scopal relations of quantifiers normally can be aligned, scopal relations can, but do not have to be built between discourse relations, and between scope-taking elements in general. Semantically, this is one of the main reasons that underspecification should be introduced rigorously. Nevertheless, some regular scopal relations may be found among discourse relations (and again in general among scope-taking elements). These relations are determined not only syntactically, but also by way of semantics and discourse structure.

The paper outlines a treatment of multiple discourse relations on the sentential level in two aspects. First, it proposes an underspecified treatment also for these cases along the lines of quantifiers and other operators. Secondly, it suggests some typical orders in which the scopal underspecification among discourse relations can be resolved. The paper is organized in the following way. In Section 2, multiple discourse phenomena are presented in terms of an example. In Section 3, the formalism of LUD is introduced. In Section 4, a representation for multiple discourse relations is proposed. Section 5 discusses possible resolutions, in which a relationship between semantics and discourse structure plays an important role.

## 2 Discourse Relations in Japanese

As mentioned above, it is apparent in Japanese that a sentence can include a number of discourse relation elements (Fig. 1). Keeping track of the assumption that all discourse relations in a sentence take a wider scope than the other scope-taking elements in a sentence, we are confronted with the next question which kind of relative scope holds among discourse relations. The treatment of discourse relations should thus be modified at least in these respects.

---

[1]Since the Verbmobil project deals with spoken languages, the unit treated is in reality not a sentence but an utterance which constitutes a turn in a dialogue and includes ellipsis and other typical phenomena which need special treatments. Here, however, the linguistically abstract unit of sentence will be presupposed.

A discourse relation is represented in LUD as a predicate with three arguments; the first one is a term for the type of the concerning discourse relation, the second one is an underspecified scope domain of the antecedent part, and the last one is another underspecified scope domain for the conclusion part. An underspecified scope domain is represented by a *hole*.

In Japanese sentences, discourse relations occur in various grammatical positions. The sentence in Fig. 1 contains at least three different discourse relations. First, there is a topic relation which is expressed by a so-called topic phrase marked by *wa*. It is encoded in the LUD as in (1) (cf. Asher's elaboration relation (Asher, 1993)). In Japanese, the antecedent part can be syntactically determined, so far as the topic phrase is expressed with the topic marker. In Fig. 1, *getsuyoubi* amounts to this part.

(1) `l2-discrel(topic,h1,h2)`

(2) `l4-discrel(explanation-noda,h5,h6)`

(3) `l3-discrel(explanation-node,h3,h4)`

Fig. 1 also contains a discourse relation expressed by the auxiliary *noda* in the modality position of the verbal complex of the conclusion part of the sentence. Semantically, it is an subordinate relation of explanation. It consists of a functional noun for the sentential nominalization *no* and the copula. The use of *noda* is different from the normal use of the copula in that it takes a temporalized sentence as a complement and, at the same time, lacks the argument of the copular predication. It is this lacking argument which makes up the conclusion part of the discourse relation (`h6` in (2)). `h5` will be bound to a DRS which is constructed out of the sentence subordinated to *noda*, that is, the whole sentence.

Finally, a discourse relation expressed by a subordinate conjunction *node* can be found in Fig. 1, too (3). This form can be seen as a participle form (*te*-form) of *noda* mentioned above. Semantically, the meaning is restricted to explanation. Therefore, the term for the discourse relation type is basically the same as (2).

Even taking these pieces of information into account, the scope relations both between *wa* and *noda* and between *wa* and *node* seem to be underspecified, whereas *noda* always has scope over *node*. Since every discourse relation has two scope domains, this observation leads to the following possibilities of scopal relations for Fig. 1.[2] These scopal relations are at least theoretically able to be forced onto the sentence in Fig. 1 (see Sec. 5).

(4) `wa(monday,noda(node(h3,h4),anaphoric))`

(5) `noda(wa(monday,node(h3,h4)),anaphoric)`

(6) `noda(node(wa(monday,h2),h4)),anaphoric)`

(7) `noda(node(h3,wa(monday,h2))),anaphoric)`

## 3 Theoretical Framework: DRT and LUD

Since the Verbmobil domain is spoken dialogues rather than isolated sentences, it is natural to choose a variant of Discourse Representation Theory, DRT (Kamp and Reyle, 1993), as the framework of its semantic formalism. To treat scope ambiguities and other underspecification phenomena adequately, we have, however, needed to extend the formalism to one which suits for representing underspecified structures (Bos, 1995). As further described in (Bos et al., 1996), LUD is a declarative description language for underspecified DRSs. The basic idea is that natural language expressions are not directly translated into DRSs, but into a representation that describes *a number of* DRSs. It is different from UDRS (Reyle, 1993) in that not only DRSs, but all predicates and discourse markers are labeled. Moreover, holes for scope domains are discerned from other labels.

A LUD-representation $U$ is a triple $U = <H_U, L_U, C_U>$, where $H_U$ is a set of holes (variables over labels), $L_U$ is a set of labeled conditions, and $C_U$ a set of constraints. Holes are special labels for the slot of an operator's scope domain. A hole will be bound by means of a plugging function to a standard label which stands for a DRS of a certain element.

The set of constraints is divided into `alfa` conditions and `leq` (*less-or-equal*) conditions. `alfa` conditions define presuppositions and anaphoric relations. They stipulate relations of those DRSs which do not come into scope relations to those DRSs which do. `leq` conditions, on the other hand, define partial order constraints between holes and labels which give a semi-lattice structure on $H_U \cup C_U$ with a hole at the top (top hole). They should be maintained in the definition of a consistent subordination relation. The latter, called a possible plugging, fully specifies the relations of holes to labels by way of an injective plugging function from holes to labels, which determines which hole is instantiated into by (or is bound to) which label. The interpretation of a possible plugging at the top hole is the interpretation of the matrix DRS. In this way, a LUD-representation describes a set of possible pluggings at once.

There are two main exceptions to this characterization of LUD. First, modifiers share its instance with the modified DRS and show no different scopal behavior. Secondly, DRSs for discourse relations are assumed to always instantiate into the top hole. In the current version, the top hole is simply assumed to be the hole argument of the sentence mood predicate of the main clause.

## 4 Representations for multiple discourse relations

In the Verbmobil semantic construction, Japanese dialogues are analysed within the same theoretical framework and with largely identical semantic macros as German ones. In order to apply the theory and implementation of LUD to Japanese, some modifications are needed. As for discourse relations, a major source of complication comes from the assumption that predicates for discourse relations have two holes as their arguments. The first problem lies in the fact that everything that goes into a `leq` relation to one hole cannot be in a `leq` relation to the other hole of the same discourse relation predicate because of its partitioning character. Another problem is the treatment of multiple occurrences of discourse relations in a sentence. We will be concentrated on the latter problem in the following sections.

For the problem of processing multiple discourse dependencies there are a few approaches (Mann et al., 1992; Kurohashi and Nagao, 1994). (Gardent, 1994) uses Tree Inserting Grammar based on the feature-based Tree Adjoining Grammar (Vijay-Shanker and Joshi, 1988) to develop a formal theory about a discourse semantic representation. This paper is distinguished from these works in two perspectives: First, it concentrates on the sentential level and offers a treatment of multiple discourse relations in terms of a formalism for underspecified structures of DRSs. Secondly, it does not concern multi-functions of one discourse relation element, but multiple occurrences of various discourse relation elements.

As suggested above, discourse relation elements have the following characteristic in LUD. The two holes which are contained in each of them partition the sentence in which the element occurs into two parts, whereas it will be subordinated to another hole by way of a `leq` constraint as a "unit". This has lead to the decision that a discourse relation element should be directly subordinated to the top hole. Other labels for DRSs should be subordinated to the discourse relation element in the way in which each of them is unambiguously subordinated to one of its two holes. The first problem mentioned at the beginning of this section can be dealt with in this manner if only one discourse relation element occurs in a sentence.

---

[2] In this example, each discourse relation element is taken as a predicate with the antecedent and the conclusion part as its arguments.

At least two problems remain when there are a number of discourse relation elements in a sentence. First, if we keep the solution above, discourse relation elements in the sentence are all candidates for the directly subordinated position to the top hole in a semi-lattice structure. Secondly, each discourse relation element introduces a different partition of the given sentence.

For a general solution, the paper proposes a device to introduce a special kind of predicate `mode` which has a hole as the only argument for the bottom of a lattice structure which is built by the top hole and discourse relation elements. This enables us to keep the decision, on the one hand, that discourse relation elements are in a next-to-top position in a possible plugging and to keep DRSs for other parts of the sentence underneath the `mode` predicate, on the other. Every discourse relation is situated above any other scope-taking element. This proposal crucially relies on the fact that for every discourse relation element which occurs in a sentence, one of its two holes can be plugged by a DRS in a lexically determined way. Additionally, it is assumed that we have a syntactic strategy in which the topic phrase is dealt with as an adjunct modification which should be interpreted in the discourse structure with respect to the main predicate of a sentence. Therefore, what is subordinated to the hole introduced by the `mode` predicate amounts to the matrix clause of the given sentence. In this way, an ordinary underspecification treatment of multiple discourse relations among each other gets possible.

For the sentence in Fig. 1, the LUD-representation can be implemented like in (8). Labels are represented under `lud_preds`. `lud_grouping` and `lud_meta` show among others which labels are to be treated together to construct DRSs. Under `lud_scoping`, `alfa` and `leq` conditions are found. The labels `l12` and `l13` are presuppositions of `l8` and `l11`. `leq` relations read that labels are always less or equal to labels in the given order. Fig. 2 is a graphical representation of the `leq` constraints of (8). Discourse relations and discourse markers are abbreviated to `discrel` and `dm`, respectively.

(8) 
```
index:       (i8,l18,h0)
lud_preds:   l1-mood(decl,h0)
             l2-discrel(topic,h1,h2)
             l3-discrel(node,h3,h4)
             l4-discrel(noda,h5,h6)
             l6-dm(i1)
             l7-predicate(getsuyoubi,i1)
             l9-dm(i2)
             l10-predicate(haitte,i2)
             l10-role(i2,arg3,i3)
             l11-role(i2,tloc,i4)
             l12-dm(i5)
             l14-dm(i6)
             l15-predicate(seminaa,i6)
             l16-dm(i7)
             l17-mode(h7)
             l19-dm(i8)
             l20-predicate(zikan,i8)
             l22-dm(i9)
             l13-neg(i9,h8)
lud_grouping:l5-inc([l6,l7])
             l8-inc([l9,l10])
             l13-inc([l14,l15])
             l18-inc([l19,l20])
             l21-inc([l22,l23])
lud_meta:    modifies(l8,l11)
lud_scoping: alfa(i6,udef,l8,l13)
             alfa(i5,pron,l11,l12)
             leq(l2,h0)
             leq(l3,h0)
             leq(l4,h0)
             leq(l5,h1)
             leq(l8,h3)
             leq(l16,h6)
             leq(l17,h2)
             leq(l17,h4)
             leq(l17,h5)
             leq(l18,h7)
             leq(l18,h8)
             leq(l21,h7)
```

The `mode` predicate can be seen as a secondary sentence mood predicate. For example, it serves in a similar way to the predicate used for the introduction of a propositional complement of propositional attitude verbs. This kind of use of the `mode` predicate does not seem to be restricted to discourse relations. For example, multiple occurences of modal expressions show a concerted behavior as regards scopal relations as in "we can perhaps meet there". The `mode` predicate is applicable when multiple occurrences of predicates in one semantic class take a scope over any other scope-taking elements together but the scope relations among each other are underspecified.

## 5 Possible Resolutions

It is sometimes possible to resolve scopal underspecifications of discourse relations on several grounds. Actually, there seems to be only one plausible resolution possibility for the sentence of Fig. 1. This resolution possibility corresponds to (5). The plugging function for this case is as follows (9). It should be read such that a label is bound to (*plugged into*) a hole.

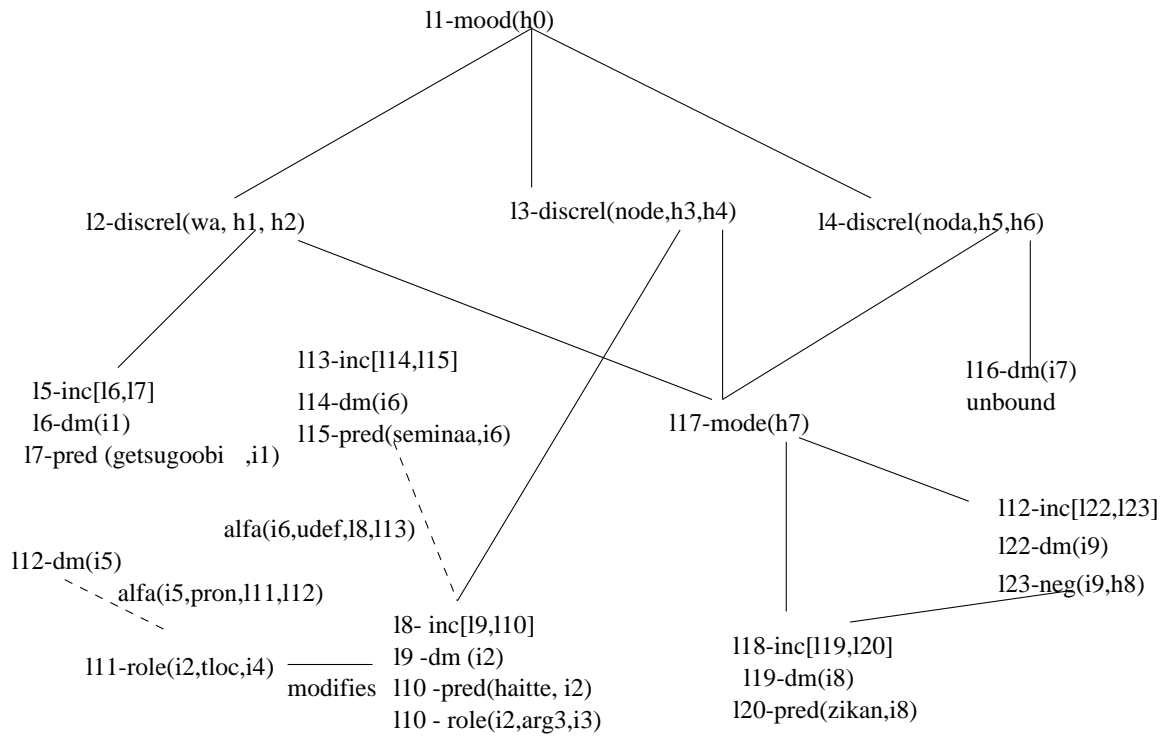

Figure 2: A graphical representation of the sentence in Figure 1

(9)    ```
       plug_into(l4,h0)
       plug_into(l2,h5)
       plug_into(l3,h2)
       ```

Confinement of resolution possibilities depends on various factors. One of the most important factors is lexical determination of the scope domains of the antecedent part or the conclusion part of a discourse relation. Especially when one of the two is determined as anaphoric, that is, sentence external, the scope of this discourse relation seems to be wider than the others. *noda* in Fig. 1 is an example for this. In the same vein, the scope of *noda* supercedes that of a conditional discourse relation *nara* in Fig. 3. The latter's scope domains of the antecedent as well as the conclusion part are sentence internal.

| gogo-nara | yamada-ga | i-ru | noda |
|---|---|---|---|
| afternoon-cond | PN-nom | be-pres | aux-pres |

*(If you mean) the afternoon, Yamada will be here*

Figure 3: Discourse relations with and without anaphoric force

Among discourse relations with sentence external anaphoric binding there are two types: those whose antecedent part is bound sentence externally and those whose conclusion part is bound sentence externally. Discourse relation particles like *dakara* (therefore) belong to the former (Fig. 4), subordinate explanation relations like *noda* belong to the latter.

| dakara | getsuyoubi-de | daijoubu-des-u |
|---|---|---|
| therefore | monday-oblwith | okay-cop-pres |

*(I) am therefore ready for monday*

Figure 4: A relation with anaphoric antecedent

Though the semantics of so-called topic phrases marked by *wa* goes beyond the scope of this paper, we assume that their discourse relations belongs to those whose antecedent part and conclusion part are both plugged sentence internally. This predicts a narrower scope than that of the subordinate relation *noda*. This not only corresponds to the intuition in (9), but is also the case in Fig. 5.

| gogo-wa | yamada-ga | i-ru | noda |
|---|---|---|---|
| afternoon-top | PN-nom | be-pres | aux-pres |

*(as for) the afternoon, Yamada will be here*

Figure 5: A topic relation getting narrow scope

On the other hand, scope underspecification among discourse relations cannot be disambiguated straightforwardly if they are of the same type according to the above classification. They can all be of the type whose antecedent and conclusion part are both bound sentence internally. In this case, the resolution seems to depend on the syntactic c-command information. This explains the stipulated scope relation between the topic *wa* and the explanative *node* in (9). (In (9), the scope relation is also influenced by antecedent resolution of the temporal-local modification which is needed from the syntactic information.) The same explanation holds for the scope difference which is observable between the two sentences in Fig. 6.

| getsuyoubi-wa | gogo-nara | daijoubu-da |
|---|---|---|
| monday-top | afternoon-cond | okay-coppres |

*As for Monday, it is ok if it is in the afternoon*

| gogo-nara | getsuyoubi-wa | daijoubu-da |
|---|---|---|
| afternoon-cond | monday-top | okay-cop-pres |

*If it is in the afternoon, the Monday is okay*

Figure 6: Topic and conditional relations

Discourse relations can, in contrast, all be of the type whose antecedent part or conclusion part is bound sentence externally. This can be observed in Fig. 7. Not only the syntactic modality auxiliary *noda*, but also the discourse particle *dakara* includes a part which is bound sentence externally. To the extent that the c-command relation is unclear between them, the resolution remains unclear here.

| dakara | ike-na-i | nodes-u |
|---|---|---|
| therefore | gomid-auxneg-pres | aux-pres |

*(It is since) (I) could not go because of it*

Figure 7: Two relations with anaphoric force

## 6 Conclusions

The LUD formalism that describes DRSs in an underspecified way also pertains to dealing with multiple discourse relation constructions, which are common in Japanese. The problem is to distinguish the discourse relations which take the wide scope relative to other scope-taking elements on the one hand and to have them underspecified among each other, on the other. The solution has a general character; several scope-taking elements can go into scope relations collectively if they belong to the same semantic class. The scope among them is underspecified again. This treatment reflects the fact that each element can introduce a different partition of the same sentence.

We have also stated an interesting semantic constraint on the resolution of multiple discourse relations which seems to prevail over the syntactic c-command constraint: discourse relations should be scopally compared with each other on the criteria whether the restriction (antecedent part) or to the scope (conclusion part) of a discourse relation has an anaphoric force.